**Corresponding Author: Samir Kumar Bandyopadhyay**, Department of Computer Science, Calcutta University, India.

**Email:** skb1@vsnl.com

# Feature Extraction of Human Lip Prints

Samir Kumar Bandyopadhyay[1], S Arunkumar[2], Saptarshi Bhattacharjee[2]

[1]Department of Computer Science, Calcutta University, India.
[2]Department of Information Technology, Institute Of Engineering & Management, India.

**Abstract:** Methods have been used for identification of human by recognizing lip prints. Human lips have a number of elevation and depressions features called lip prints and examination of lip prints is referred to as cheiloscopy. Lip prints of each human being are unique in nature like many others features of human. In this paper lip print is first smoothened using a Gaussian Filter. Next Sobel Edge Detector and Canny Edge Detector are used to detect the vertical and horizontal groove pattern in the lip. This method of identification will be useful both in criminal forensics and personal identification. It is our assumption that study of lip prints and their types are well connected to play a song in a better way that are well accepted to people who loves to hear songs.

Introduction:

Human identification from the study of their biometrics has gained much popularity in recent times. In these approaches human beings can be identified based on their physical traits without the aid of any external key. From the science of cheiloscopy recognition of human beings is originated based on the study of their lip prints [2]. Several methods are used for human identification such as face, iris, retina, finger veins, skin, finger-nails recognition, palm vein, etc. [6-9].
 Yasuo Tsuchihasi and Kazuo Suzuki [4-5] at Tokyo University from 1968 to1971 confirmed that humans may have unique lip features in general. These examinations helped scientists for recognition of human based on lip prints.
Dental, fingerprint, palm vein, iris, finger prints and DNA comparisons are the most common methods used for human identification processes. Sometimes it is not possible to use these methods due to some unknown reasons then lip prints can be used in place of these approaches. Lip prints are also used to support the sex determination of the examined subject [3].

The biological features of human are the first to be noted in 1902 by anthropologists, R. Fischer [10]. Edmond Locard [11] is first recommended by Personal identification and criminalization using lip prints.

Sometimes lip print can be a basis for crime detection. It is used to find the situation on the basis of evidence surrounding the crime spot for identifying number of people involved, their nature, sex as well as type of crime held during the event.

Research studies and information regarding the use of lip prints as evidence in personal identification and criminal investigation are very much necessary. This paper aims to study the area and proposes a method for detecting grooves in the lip of the human using existing edge detection methods.

### REVIEW WORK

1. Lip print characteristics have been widely used in forensics by experts and in criminal police practice for human identification. The grooves in the human lips are unique to each person and are used to determine human identification.
2. While examining human lips characteristics the anatomical patterns on the lips are taken into account. The pioneer of Chieloscopy, Professor J. Kasprzak, used 23 lip patterns [1-2] for finding features of human being. Such patterns (lines, bifurcations, bridges, pentagons, dots, lakes, crossings, triangles etc.) are very similar to fingerprint, iris or palm print patterns. The statistical characteristics features extracted from the lip prints also account for unique identification.
3. This paper concentrates on studying both statistical and anatomical data in order to successfully identify a particular lip print. Initially, this paper geometrically determines the various statistical data (like upper to lower lip height ratio, upper lip height to width ratio). Based on these proposed method performs the preliminary verification. Anatomical verification deals with studying the various physical indentations and groove patterns in the human lip print and matching them with the print available in the database. The given image is considered to be on a white background. The lip print is first smoothened using Gaussian Filter to remove the noise that has crept in during the image capture. Sobel edge detector and Canny edge detector are used to detect the groove edges in the lip. It must be noted that Sobel operator might be beneficial in this case since Canny edge detector detects the weak edges also unlike Sobel, which concentrates on only the most prominent changes. Therefore, a higher accuracy can be achieved by the Sobel operator over the Canny operator.

### BACKGROUND WORKS

Edge detection is a problem of fundamental importance in image analysis. In a typical image, edges Identify object boundaries and are therefore useful for segmentation, registration, and identification of objects in a scene.

An edge is the variations of gray level values in boundary between an object and the background. The shape of edges in images depends on the geometrical and optical properties of the object, the illumination conditions, and the noise level in the images.

In practice, sampling, and other image acquisition imperfections yield edges that are blurred, with the degree of blurring being determined by the factors such as quality of the image acquisition system, the sampling rate, and illumination conditions under which the image is acquired. As a result, edges are more closely modeled as having a "ramp like" profile. The slope of the ramp is inversely proportional to the degree of blurring in the edge. The blurred edges tend to be thick and sharp edges tend to be thin.

The Sobel Operator is used for edge detection in the field of Image Processing. Its operation is based on the principle of discrete differentiation of first order. The operation yields the approximate abso-lute value of the gradient of the image intensity function at each and every point of the image. It accentuates the regions of high spatial gradient that corresponds to edges. The Canny operator is an optimal edge detector, which accepts a gray scale image as an input and shows the positions of tracked intensity discontinuities as an output. It employs a multi-stage algorithm, in which the raw image is convolved with a Gaussian filter. The output is a blurred version of the original which is not affected by a much small pixels.

Non – maximal suppression technique is applied to detect whether the gradient magnitude assumes local maximum in the direction of the gradient; therefore at the end of this search, a set of edge points is obtained as a binary image. Canny operator uses thresholding with hysteresis to detect whether a point in this binary image corresponds to an edge or not. This process requires a high threshold and a low threshold. A point having a high gradient corresponds to an edge point; therefore high threshold is applied initially. Thereafter, low threshold value determines the finer detail points of the image. The final image gives us sufficient detail to detect edge points from non-edge points.

## PROPOSED METHOD

1. Pre - Processing is done on the image to reduce noise. The given image (Fig 1(a)) is first converted into gray scale. The average gray level value is calculated and subsequent comparison is done with each pixel to determine object vis-a-vis background pixels. A special one-dimensional case of the *k-means clustering* iterative algorithm, which has been proven to converge at a *local* minimum, is used to calculate the average gray level value. Background pixels of the image is made black (that is zero intensity) keeping the pixels of the lip print (object) intact (Fig 1 (b)).
2. Algorithm to compute Average Gray Level (special case of iterative k - means clustering algorithm):
3. Fast Fourier Transform is applied on image in (Fig 1(b)). The first element of the transformed matrix i.e. row index =1 and column index=1 contains the sum of all the intensity values. This value when divided by the total number of pixels in the image gives the average intensity value. This is assumed to be the initial threshold value $t_i$.
4. The image is segmented into object and background pixels, thus creating two sets:

$$G_1 = \{f(m,n):f(m,n)< t_i\} \text{ (object pixels)}$$
$$G_2 = \{f(m,n):f(m,n)>= t_i\} \text{ (background pixels)}$$

The average of each set is computed as:

$$m_1 = \text{average value of } G_1$$
$$m_2 = \text{average value of } G_2$$

A new threshold is created that is the average of $m_1$ and $m_2$:

$$t_f = (m_1 + m_2)/2$$

Repeat steps 2 -4 until the new threshold ($t_f$) converges to $t_i$.

$$|t_i - t_f| <= e$$

In this case the error *e is considered to be 1 since intensity values of pixels are integers*. The transformed image after thresholding is then smoothened using Gaussian smoothing filters.

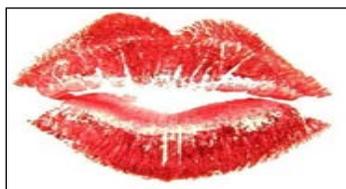

Fig 1(a): The given image

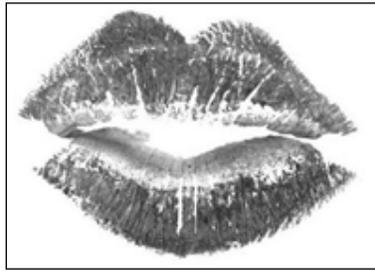

Fig 1(b): The image after conversion into gray scale

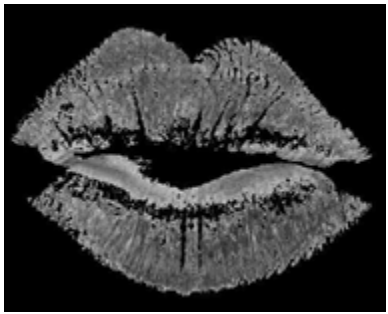

Fig 1(c): The image after thresholding

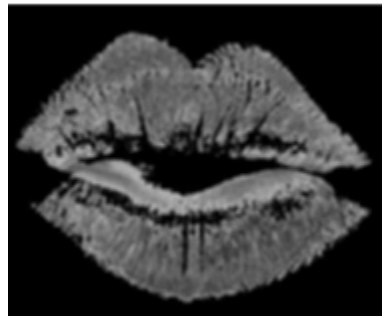

Fig 1(d): The image after passing through 7X7 Gaussian Filter

Smoothing is done repeatedly four times to reduce all the noise in the image and consider only the significant grooves of the lip prints. Small edges or grooves which are highly insignificant get lost (Fig 1(d)).

In the next level the primary grooves of the lip prints are detected in two phases:

<div style="text-align:center">Phase I: Vertical groove detection<br>Phase II: Horizontal groove detection</div>

**Edge Detection**

Before detecting the edges or the groves the image is passed through a Gaussian filter four times. The smoothing effect on the image at subsequent levels is shown in Fig (A) Fig (B) Fig (C) and Fig (D).

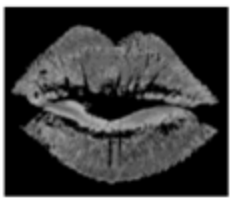 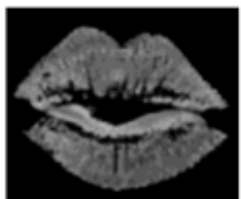 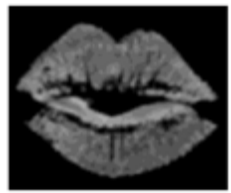 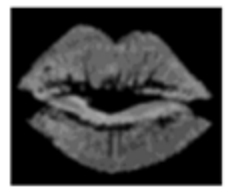

Fig: (A)   Fig: (B)   Fig: (C)   Fig: (D)

Edges or grooves are detected horizontally as well as vertically using horizontal as well as vertical Sobel operators. It is seen that only prominent edges are detected in these operations (Fig 1 (e) and Fig 1(f)). However the shape of the lip print or the lip contour becomes a bit distorted and still less significant grooves are detected. Hence the images are again smoothed using a Gaussian filter and respective edge operations are done again on these images (Fig1 (g) and Fig1 (h)). The images Fig1(g) and Fig 1(h) shows that the detection of lesser significant groves has decreased but still the image shape is not restored. The image shape or contour is restored by passing it again through a Sobel Edge Detector (Horizontal as well as Vertical) Fig 1(i) and Fig 1(j).

Canny Edge Detection is done on the complement of image (Fig 1(k) and Fig 1(l)) and the primary grooves are thus detected efficiently. The final image is shown in Fig (m) and Fig (n).

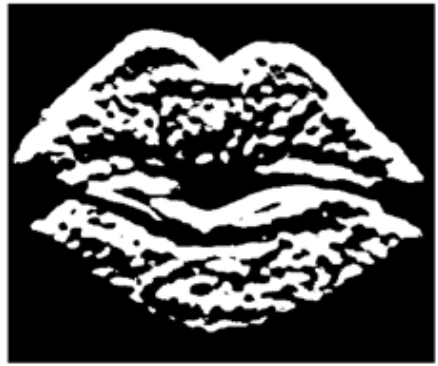

Fig 1(e): Horizontal Sobel Operator applied on Fig 1(d)

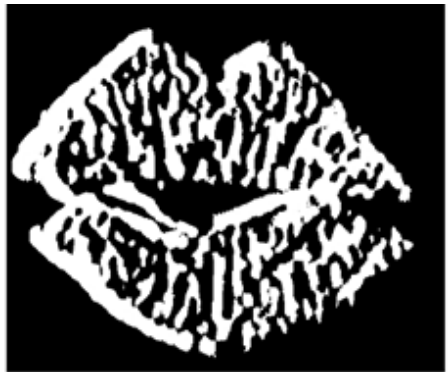

Fig 1(f): Vertical Sobel Operator applied on Fig1(d)

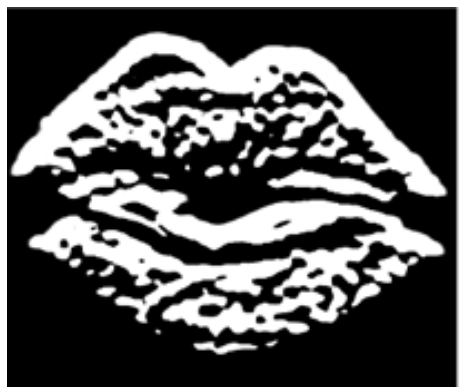

Fig 1(g): Gaussian smoothing on Fig 1(e)

SK Bandyopadhyay, Journal of Current Computer Science and Technology, 2 (1), 2012, 01-08

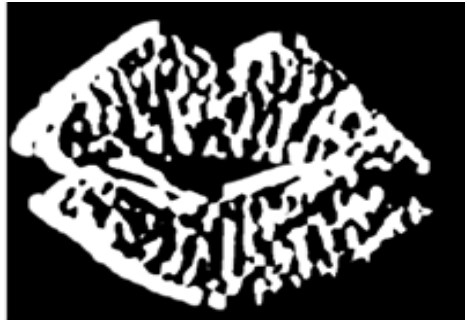

Fig1(h): Gaussian smoothing on Fig 1(f)

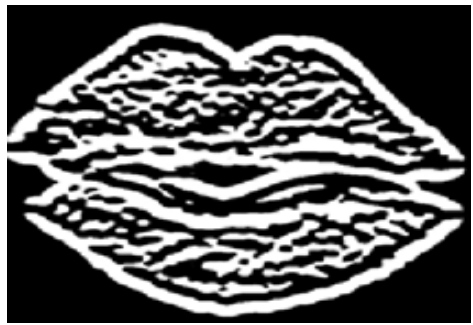

Fig 1(i): Horizontal Sobel Operator on Fig 1(g)

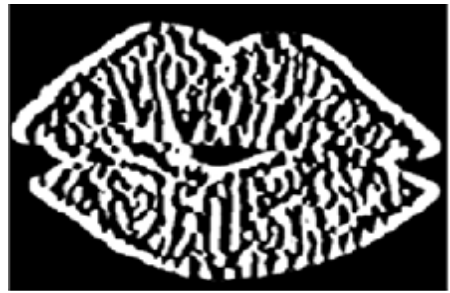

Fig 1(j): Vertical Sobel operator on Fig 1(f)

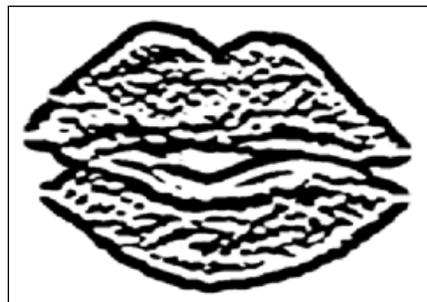

Fig 1(k): Complement of Fig 1(i)

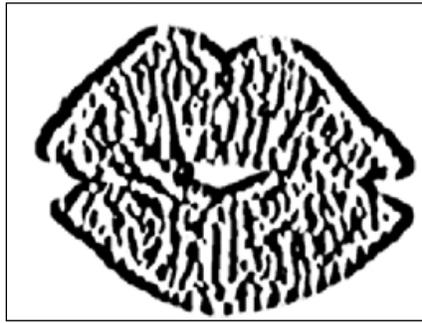

Fig 1(l): Complement of Fig 1(j)

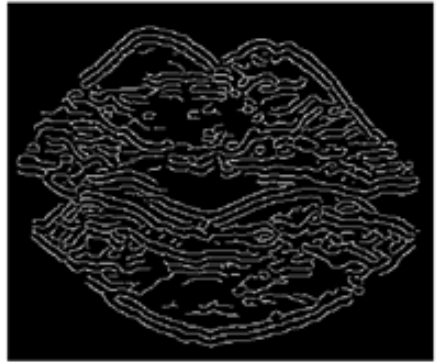

Fig 1(m): Applying canny edge detector on Fig 1(k)

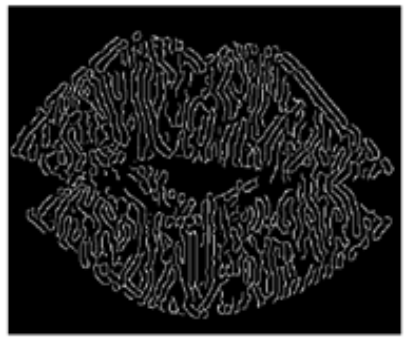

Fig 1(n): Applying canny edge detector on Fig 1(l)

Results are obtained quiet satisfactory. The aforementioned algorithm is used to extract the feature details in a human lip. The final image obtained can be used to identify human beings uniquely by matching it with an existing lip print.

**Conclusion**

This paper is based on studying both statistical and anatomical data in order to successfully identify a particular lip print. The lip print is first smoothened using Gaussian Filter to remove the noise that has crept in during the image capture. Sobel edge detector and Canny edge detector are used to detect the groove edges in the lip. The proposed method has achieved promising recognition results for well detected lips images and it motivates us to recognize person based on lip.